\documentclass[conference]{IEEEtran}
\IEEEoverridecommandlockouts
\usepackage{cite}
\usepackage{amsmath,amssymb,amsfonts}
\usepackage{algorithmic}
\usepackage{graphicx}
\usepackage{textcomp}
\usepackage{xcolor}

\usepackage{makecell}

\def\BibTeX{{\rm B\kern-.05em{\sc i\kern-.025em b}\kern-.08em
    T\kern-.1667em\lower.7ex\hbox{E}\kern-.125emX}}
\begin{document}

\title{Automated Assessment of Multimodal Answer Sheets in the STEM domain\\

}

\author{
\IEEEauthorblockN{Rajlaxmi Patil}
\IEEEauthorblockA{\textit{Computer Engineering Department} \\
\textit{Pune Institute of Computer Technology}\\
Pune, Maharashtra \\}
\and
\IEEEauthorblockN{Aditya Ashutosh Kulkarni}
\IEEEauthorblockA{\textit{Computer Engineering Department} \\
\textit{Pune Institute of Computer Technology}\\
Pune, Maharashtra \\}
\and
\IEEEauthorblockN{Ruturaj Ghatage}
\IEEEauthorblockA{\textit{Computer Engineering Department} \\
\textit{Pune Institute of Computer Technology}\\
Pune, Maharashtra \\}
\and
\IEEEauthorblockN{Sharvi Endait}
\IEEEauthorblockA{\textit{Computer Engineering Department} \\
\textit{Pune Institute of Computer Technology}\\
Pune, Maharashtra \\}
\and
\IEEEauthorblockN{Dr. Geetanjali Kale}
\IEEEauthorblockA{\textit{Computer Engineering Department} \\
\textit{Pune Institute of Computer Technology}\\
Pune, Maharashtra \\}
\and
\IEEEauthorblockN{Raviraj Joshi}
\IEEEauthorblockA{\textit{L3Cube Labs} \\
\textit{Indian Institute of Technology, Madras}\\
Pune, Maharashtra \\}
}

\maketitle

\begin{abstract}
In the domain of education, the integration of technology has led to a transformative era, reshaping traditional learning paradigms. Central to this evolution is the automation of grading processes, particularly within the STEM domain encompassing Science, Technology, Engineering, and Mathematics. While efforts to automate grading have been made in subjects like Literature, the multifaceted nature of STEM assessments presents unique challenges, ranging from quantitative analysis to the interpretation of handwritten diagrams. To address these challenges, this research endeavors to develop efficient and reliable grading methods through the implementation of automated assessment techniques using Artificial Intelligence (AI). Our contributions lie in two key areas: firstly, the development of a robust system for evaluating textual answers in STEM, leveraging sample answers for precise comparison and grading, enabled by advanced algorithms and natural language processing techniques. Secondly, a focus on enhancing diagram evaluation, particularly flowcharts, within the STEM context, by transforming diagrams into textual representations for nuanced assessment using a Large Language Model (LLM). By bridging the gap between visual representation and semantic meaning, our approach ensures accurate evaluation while minimizing manual intervention. Through the integration of models such as CRAFT for text extraction and YoloV5 for object detection, coupled with LLMs like Mistral-7B for textual evaluation, our methodology facilitates comprehensive assessment of multimodal answer sheets. This paper provides a detailed account of our methodology, challenges encountered, results, and implications, emphasizing the potential of AI-driven approaches in revolutionizing grading practices in STEM education.

\end{abstract}


\section{Introduction}
In the current digital age, the field of education is undergoing a profound transformation, coupled with advancements in technology. From virtual laboratories to interactive learning platforms, the integration of technology has revolutionized the way students engage with content and lessons taught. However, the area that has garnered significant attention is automating the process of grading assessments in the STEM domain. STEM encompasses the subjects of Science, Technology, Engineering, Medical. While there have been attempts in multiple languages to grade and assess subjects predominantly of written work, such as Literature \cite{c5}. Grading STEM subjects offers several challenges to grade, such as quantitative analysis,  complex mathematical concepts, a wide range of formats, short answer responses, code snippets, and the presence of ambiguity and subjectivity in evaluating certain responses, especially in open-ended questions. Moreover, most importantly, STEM assessments involve handwritten diagrams and sketches which pose challenges for systems in terms of accurate recognition and interpretation. Ensuring robust handwriting recognition and diagram understanding capabilities is crucial for reliable grading. This necessitates effective evaluation methods that accurately measure students understanding and proficiency in STEM.
This research work aims to address the need for efficient and reliable grading methods in the STEM domain through the development of automated assessment techniques using Artificial intelligence.
Our contributions can be summarised as follows:
\begin{enumerate}
    \item \textbf{Efficient Textual Answer Evaluation in STEM}: Our first contribution lies in the development of a robust system for evaluating textual answers in the STEM domain. By incorporating a sample answer as a reference point, our system excels in identifying and assessing the nuances of student responses with high efficiency. Through advanced algorithms and natural language processing techniques, we enable precise comparison and grading, providing educators with valuable insights into students' comprehension and proficiency levels.
    \item \textbf{Context-Aware Diagram Evaluation, Emphasising Flowcharts}: Our second contribution focuses on enhancing the evaluation of diagrams, particularly flowcharts, within the STEM context. Recognizing the importance of understanding context in diagram assessment, our system employs sophisticated techniques to analyze and interpret flowcharts efficiently. By considering the logical flow and coherence of diagrams about the given context, our system ensures accurate evaluation while minimizing manual intervention.
\end{enumerate}
Addressing the challenges of grading STEM assessments using automated systems is critical for enhancing the efficiency, accuracy, and fairness of evaluating practices in STEM education. 
In the subsequent sections of this paper, we will provide a detailed account of our proposed methodology for evaluating textual answers and flowchart-based diagrams in STEM. We’ll also be going through the challenges we encountered, the results, and the implications of our research in the field of artificial intelligence. Our goal extends beyond the immediate goal of evaluation of flowchart-based diagrams. By addressing the challenges, we hope to inspire similar efforts for other types of diagrams, and edge cases and demonstrate the potential for multi domain evaluation using AI.

\section{Related Work}

In recent years, advancements in evaluation techniques for grading answers and diagrams have led to a broader range of ways to correct answers beyond simple keyword matching. Both keywords and contextual understanding now play crucial roles in the assessment of textual answers, and in the evaluation of flowchart diagrams, both structure, the relation between entities, and context play an important role.
Moreover, there have been several notable datasets that have been made accessible for researchers, such as (Montellano et al. (2022)) \cite{c2} which have contributed to a valuable dataset comprising 775 handwritten flowchart images in Spanish and English. Moreover, ( Bresler et al., 2016)\cite{c3} Bresler and his collaborators from the Czech Technical University. (Shukla et al.2023) \cite{c4} introduces the FloCo dataset, which contains 11.8K images of flowcharts along with their corresponding Python codes. The dataset includes both synthetic and handwritten images, further enriching the available resources for flowchart-scoring research
present a dataset encompassing 300 finite automata diagrams and 672 flowchart diagrams.

Regarding evaluation methods for textual answers and structures for evaluating diagrams,
Olowolayemo et al. (2018) \cite{c6} introduced evaluation methods such as Levenshtein distance and cosine similarity to measure the similarity between reference and actual answers. Cosine similarity, in particular, has shown high agreement with human graders. Cosine similarity between two texts can be computed as the cosine of the angle between the vectors.
Moreover, there is a study that proposes using a bidirectional encoder representation from the transformer BERT for answer grading systems. Pre-trained BERT overcomes the problem of small training datasets. 
Lubis et al. proposed a BERT-based deep neural network framework for answer grading, utilizing word embeddings based on semantic similarities between students’ answers and the reference answer and syntactic analysis to achieve a 70\% similarity with manual grading. This process also uses POS tagging, syntactic analysis, and dependency parsing. The model calculates sentences containing negations between student’s responses and links with the answer.
In image analysis, various papers have contributed significantly. Russakoff et al. (2004)\cite{c7} introduced Regional Mutual Information for image similarity that takes spatial information into account. It finds applications in both 2-D and 3-D applications in fields of X-ray analysis. This helps to measure image similarity, which can be valuable in various medical contexts. Ding et al. (2022) \cite{c10} presented DISTS for Image Quality Assessment, leveraging texture and structure similarity suitable for image processing tasks, including denoising, deblurring, super-resolution, and compression. It can predict human quality ratings for natural photographs.
Julca et al. (2020) proposed a graph grammar approach for online handwritten graphics such as math expressions and flowchart recognition using graphical parsing, that parses diagrams from top to bottom. Yun et al. (2022) introduced Instance GNN for complex symbol recognition.
Text line recognition saw an innovation by Bhunia et al. (2021) \cite{c9}, focusing on OCR applications, which focus on text line recognition, a crucial component of OCR. Additionally, Praveen et al. (2014) introduced a model for automatic evaluation of subjective questions in online exams. The paper discusses methods employed in the system, future work, WordNet, and similarity measures like string match. The system aims to incorporate the answers of students into a graphical representation and it can apply similarity measures such as string match.
Noorbehbahani and Kardan modified the BLEU algorithm into M-BLEU for assessing free text responses, showing a strong correlation with human scores. Challenges include maintaining diverse reference answers and scalability.
Finally, Fang et al. (2022) \cite{c26} introduced DrawnNet, a CNN-based keypoint detector for hand-drawn diagrams, extending the CornerNet model. This study extends the CornerNet model with the arrow orientation prediction branch, which makes the ability to detect key elements better.
These advancements signify a shift towards more nuanced evaluation techniques across various domains, including text and image analysis, aiming for more accurate and efficient assessment methods

\section{Methodology}
In this section, we detail the methodology employed for the automated assessment of multimodal answer sheets. Our approach consists of two key stages:
Textual Evaluation and Diagram Evaluation. The sample answer sheets are the answer sheets that have to be evaluated against the model or sample answer key, which acts as a baseline for comparative analysis and grading. 

The image is first segmented into various blocks containing textual data and diagrams. These blocks are classified as diagrams based on the text area density of the image. For text extraction, CRAFT is used to identify text regions in the image. Based on affinity score and region score a score map is generated by the CRAFT model. From this score map dimensions and the location of individual characters can be obtained. This is further used to combine multiple characters belonging to a word into a single instance. In this manner coordinates of bounding boxes for each word are generated. This is used for line segmentation of text. Each line is passed to TROCR \cite{c22}for optical character recognition. Text for each answer is extracted and stored along with its corresponding diagrams(if present) in a mapped data structure. As TrOCR is a transformer-based model it works better with line segmentation instead of word segmentation as extra context is present for text extraction in the case of line segmentation.

Diagram evaluation is a crucial aspect of the answer sheet assessment process, primarily due to the inherent variability in how diagrams can effectively convey information. It is insufficient to merely compare two diagrams based on their visual appearance; structural differences can exist between diagrams that both accurately represent the intended concepts. Consider, for instance, flowcharts used in problem-solving contexts: one student may label decision blocks with "yes/no," while another may opt for "true/false" in the same position. Although the textual representation differs, the contextual meaning remains consistent. Hence, diagram evaluation necessitates a deeper level of understanding and context awareness.
To address this challenge, our approach involves transforming diagrams into textual representations. By encoding the visual information into a textual format, we enable a more nuanced assessment process that leverages semantic understanding rather than solely visual comparison. Subsequently, we employ a Large Language Model (LLM) for inference, enabling the system to impart the necessary contextual knowledge and assign the most appropriate score to each input diagram.

\subsection{Feature extraction}\label{AA}
The data extraction process is the first step of the evaluation pipeline. The data extraction process: Initially each page of the answer sheet is divided into several blocks. The image is converted to grayscale. Then vertical margins and horizontal lines present on answer sheets are removed. Contour detection is used to separate the diagram region and text regions. CRAFT text detector is used to identify text regions in these blocks. CRAFT provides bounding boxes for text present in the blocks. Using the coordinates of the bounding boxes, the image is segmented line by line, and any blank space present at the beginning or end of the line is removed to prevent hallucinations from the TrOCR model. Coordinates of the bounding boxes are used to calculate the ratio of the area of the textual region to the area of the image. This ratio is used to determine if the block contains answer text or diagram. Each of these lines is then passed to TrOCR for text extraction. Text extracted from each block is stored in a dictionary corresponding to the question it belongs to along with the diagram. 

\

% \item CRAFT: CRAFT is a model by Clova AI research, Naver Corp. It provides scene text detection. It performs word-by-word segmentation for text present in flowchart boxes. Each word and mathematical symbol for each block in the flowchart is detected using EasyOCR and stored for the corresponding block of the flowchart. 

% \item Yolo V5: YoloV5 is an object detection model algorithm developed by Ultralytics. This model is based on the You Only Look Once family of computer vision models. This model mainly consists of 3 components.
%     \begin{itemize}
%         \item The main body of the network is called the Backbone, which is designed using the CSP-Darknet53 structure. In this layer, the process of feature extraction is performed. Both low-level features and high-level features are extracted from the given image layers.
%         \item The second component named Neck, is the connecting link between the backbone and the head. This component plays a role in integrating the positional features with the semantic features extracted from the image layers by the backbone. 
%         \item The final component, called the Head, is key in detecting the objects present in the images. Detected images may be of varied sizes and are output as a vector. The output vector contains all the information related to the image ranging from its probability to its bounding box-related information.
%     \end{itemize}

\subsection{Model evaluation}

\subsubsection{Evaluation of Diagrams}
    Diagrams are passed to the YoloV5 model from coordinates of various blocks and arrows present in the diagram are obtained. Duplicate blocks are removed. Later, using arrow and arrowhead coordinates neighbors of each block are found. Information about each component in the diagram is obtained in this way. Using this a textual representation of a flowchart is obtained. The textual representation of each block consists of: 
    \begin{enumerate}
        \item Block ID
        \item Number of its neighbors
        \item The previous block in the flowchart
        \item Next block or blocks in case of conditional statements
        \item Block type: Start/Stop/Condition/Process
        \item Text present in the block
    \end{enumerate}
    This is passed to LLM along with the model answer diagram’s representation for evaluation. \\

\subsubsection{Evaluation of Textual Answers}
The textual evaluation process consists of passing the input answer and the model answer to an LLM with a prompt. The prompt tells the LLM to evaluate the input answer by comparing it to the model answer in accordance with certain factors. The factors include grammatical correctness, the structure of the sentence, and important points coverage. The system is designed in such a way that any LLM can be plugged in easily. The system currently uses MistralAI which is an open source LLM. The LLM evaluates the input answer and generates a score which is question-wise segmented.

\section{Methodology}
\subsection{CRAFT (Character region Awareness for Text Detection)}
CRAFT is a CNN model based on VGG-16. The output is created by combining two channels to generate score maps. This is done using affinity score and region score. The affinity score is obtained by exploring each character and the affinity between characters. Region score is used for character segmentation. These characters are later grouped into a single word using the affinity score. The CRAFT model outperforms state-of-the-art models on benchmark datasets \cite{c21}.

\subsection{TrOCR(Transformer-based Optical Character Recognition)}
It is an encoder-decoder model using a pre-trained image Transformer model as an encoder and a pre-trained text Transformer as a decoder. It was trained on the IAM Handwriting Database and on synthesized handwritten text line data generated using 5427 handwritten fonts. TrOCR achieves state-of-the-art results on printed as well as handwritten text \cite{c22}.
\subsection{YOLO (You Only Look Once)}
YoloV5 is an object detection model algorithm developed by Ultralytics. This model is based on the You Only Look Once(YOLO)\cite{c20} family of computer vision models. This model mainly consists of 3 components. 
\begin{enumerate}
    \item The main body of the network is called the Backbone, which is designed using the CSP-Darknet53 structure. In this layer, the process of feature extraction is performed. Both low-level features and high-level features are extracted from the given image layers.
    \item The second component named Neck, is the connecting link between the backbone and the head. This component plays a role in integrating the positional features with the semantic features extracted from the image layers by the backbone. 
    \item The final component, called the Head, is key in detecting the objects present in the images. Detected images may be of varied sizes and are output as a vector. The output vector contains all the information related to the image ranging from its probability to its bounding box-related information.
\end{enumerate}
\subsection{Azure OCR}
It is a component of Azure AI vision. It uses deep learning-based models to extract text from images. It works on handwritten and printed text present on different backgrounds like text on whiteboards, receipts, posters, or invoices. Azure was used to extract various steps and mathematical expressions present in flowcharts.
\subsection{FastAPI}
It is a high-performance web framework used to create RESTful APIs in Python. It is built on top of the Starlette web server. It contains features like automatic error handling and data validation.
\subsection{Mistral AI }
Mistral-7B, developed by TheBloke, stands as a colossal language model with around 7 billion parameters, placing it among the largest models in natural language processing (NLP). Leveraging transformer architecture, it boasts versatility across a spectrum of NLP tasks including text completion, summarization, translation, and sentiment analysis. Pre-trained on extensive textual data, Mistral-7B captures intricate language patterns and semantics. Its adaptability allows for fine-tuning on specific tasks or domains, empowering developers to tailor its capabilities. Widely applicable, Mistral-7B finds use in diverse applications such as chatbots, question-answering systems, and document analysis, consistently delivering state-of-the-art performance across various benchmarks \cite{c25}.

\begin{figure}[h]
        \begin{center}
		\includegraphics[scale=0.22]{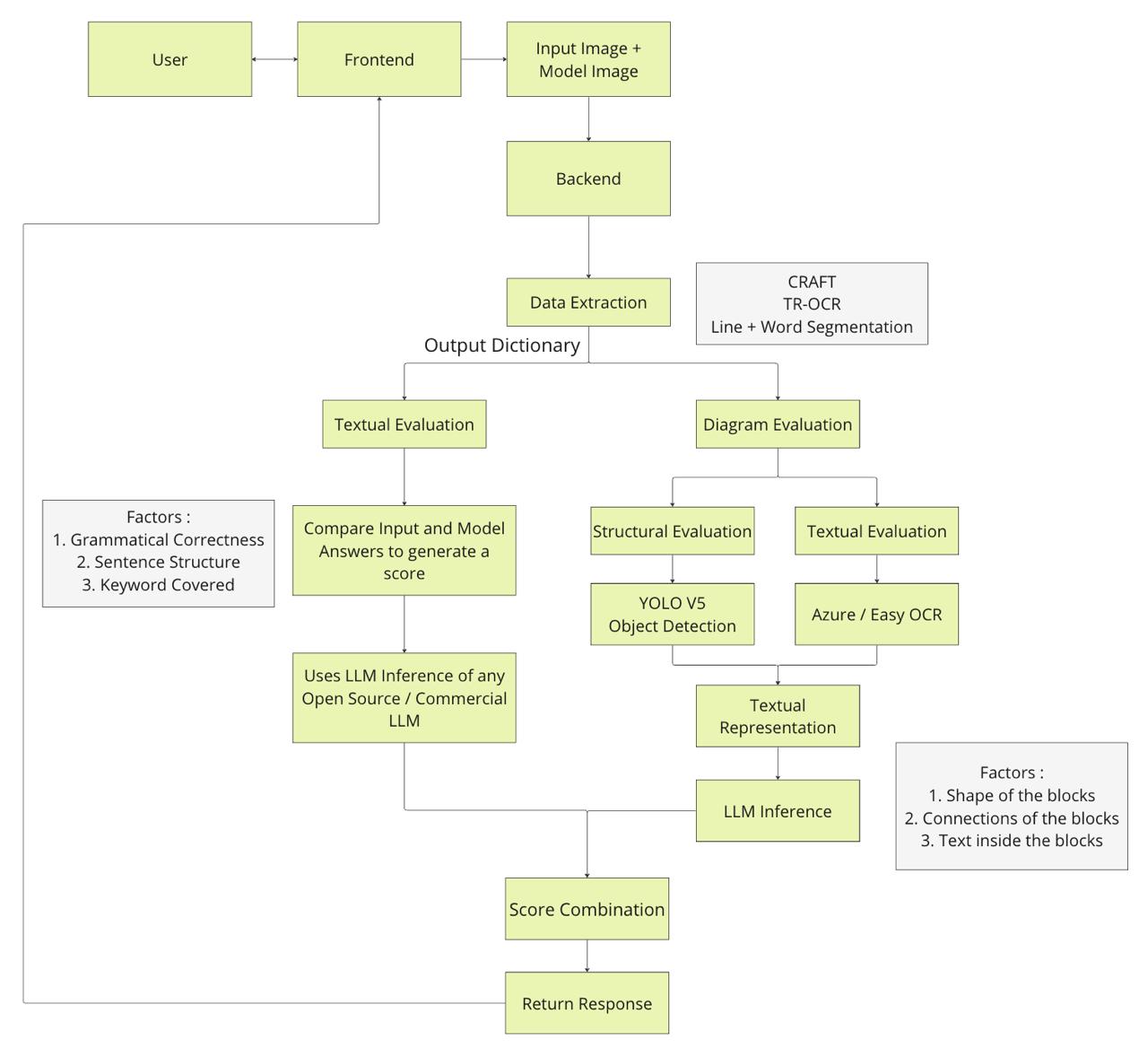}
            \caption{System Architecture Diagram}
            \label{fig:enter-label}
	\end{center}
\end{figure}

% Number equations consecutively. To make your 
% equations more compact, you may use the solidus (~/~), the exp function, or 
% appropriate exponents. Italicize Roman symbols for quantities and variables, 
% but not Greek symbols. Use a long dash rather than a hyphen for a minus 
% sign. Punctuate equations with commas or periods when they are part of a 
% sentence, as in:
% \begin{equation}
% a+b=\gamma\label{eq}
% \end{equation}

% Be sure that the 
% symbols in your equation have been defined before or immediately following 
% the equation. Use ``\eqref{eq}'', not ``Eq.~\eqref{eq}'' or ``equation \eqref{eq}'', except at 
% the beginning of a sentence: ``Equation \eqref{eq} is . . .''
\section{Algorithm}

\begin{enumerate}
        \item We give the input image and the model image to the model.
        \item There is a data extraction process that takes place here, which uses TR-OCR and a process of line-by-line and word-by-word segmentation to obtain an input dictionary that is question-wise segmented
        \item The process is then divided into 2 flows, textual evaluation and diagram evaluation.
        \item The textual evaluation uses an LLM to compare the input answer and the associated model answer in terms of factors such as grammatical correctness, structure of the sentence, and important points covered or not and accordingly assigns a score to the input answer.
        \item The diagram evaluation is divided into 2 parts : 
        \begin{enumerate}[a]
            \item Structural Evaluation: This uses a YOLO v5 object detection model and CNNs to understand the structure and the flow of the diagram. After this we obtain the shape of the blocks, the connection of blocks, and their order.
            \item Textual Evaluation: This uses Azure OCR / Easy OCR to extract the data present inside the blocks.
        \end{enumerate}
        \item After these steps the diagram is converted into a textual representation. 
        \item The textual representation of both the model diagram and the input diagram is passed to LLM for inference and it is judged based on factors such as: Shape of the blocks, the connection of the blocks and their order, and the text present inside the blocks. The LLM imparts knowledge here which is a key factor in the evaluation and generates a score for the input diagram.
        \item The textual and diagram scores are combined to generate a final response which contains all the scores segmented in a question-by-question manner.
    \end{enumerate}
    \begin{figure}[h]
        \begin{center}
		\includegraphics[scale=0.4]{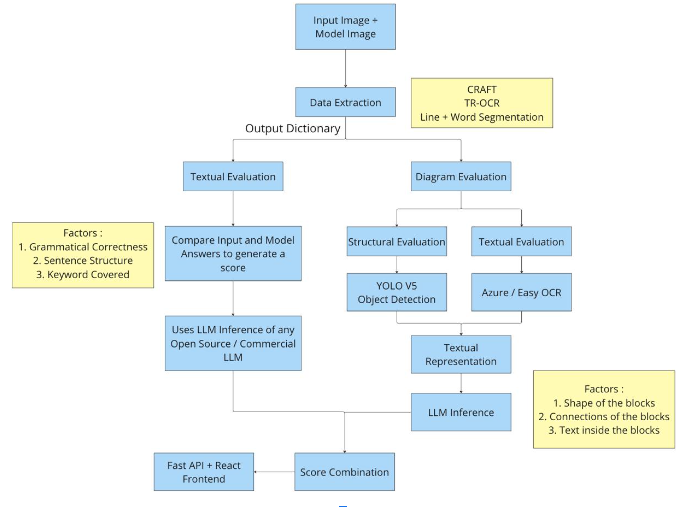}
            \caption{Algorithmic Flow Diagram}
            \label{fig:enter-label}
	\end{center}
    \end{figure}

\section{IMPLEMENTATION CHALLENGES AND SOLUTIONS}
\subsection{OCR Challenge}
One of the major challenges faced during the implementation was OCR. The first step of evaluation is data extraction and extraction of handwritten text which might be cursive is not a straightforward process as the accuracy of that with predominantly utilized OCRs like tesseract OCR has a very low accuracy. To overcome this, TrOCR was used which works well with handwritten text. However, TrOCR when used as is was not giving very good results due to various factors affecting its accuracy. To overcome this, a line-by-line segmentation process was followed after removing any extra spaces in a line and then the line was passed to TrOCR for character recognition, which improved the accuracy. 

\subsection{Diagram Discrepancy}\label{SCM}
If we are comparing two diagrams as is there are several many in which the result can have a lower score than expected. For instance, when flowcharts are considered, the decision blocks are to be evaluated, the text inside can be true/false or it can be yes/no. Both contextually represent the same thing but visually are different. So, if two images are directly compared based on their visual similarity then that is bound to give lower results. This limitation was effectively addressed by creating a diagrammatic representation of the diagrams and then passing those representations to an LLM. The LLM then uses knowledge to overcome these problems and gives more accurate results. 

\section{RESULTS AND DISCUSSION}
Following are the test cases that we have considered for evaluation:
\begin{enumerate}
    \item Textual Evaluation: 
    \begin{itemize}
        \item This involves a separate evaluation of textual answers data which is compared with the sample answer key, and evaluated.
        \item For testing the textual evaluation module we manually compared
        the answers given by students with the ideal answer and checked
        whether the justification and score given for the evaluation were
        correct.
        \item The system was able to identify the keywords and concepts present
        in answers and accordingly, generate scores.
    \end{itemize}
    \begin{figure}[h]
        \centering
        \includegraphics[scale=0.18]{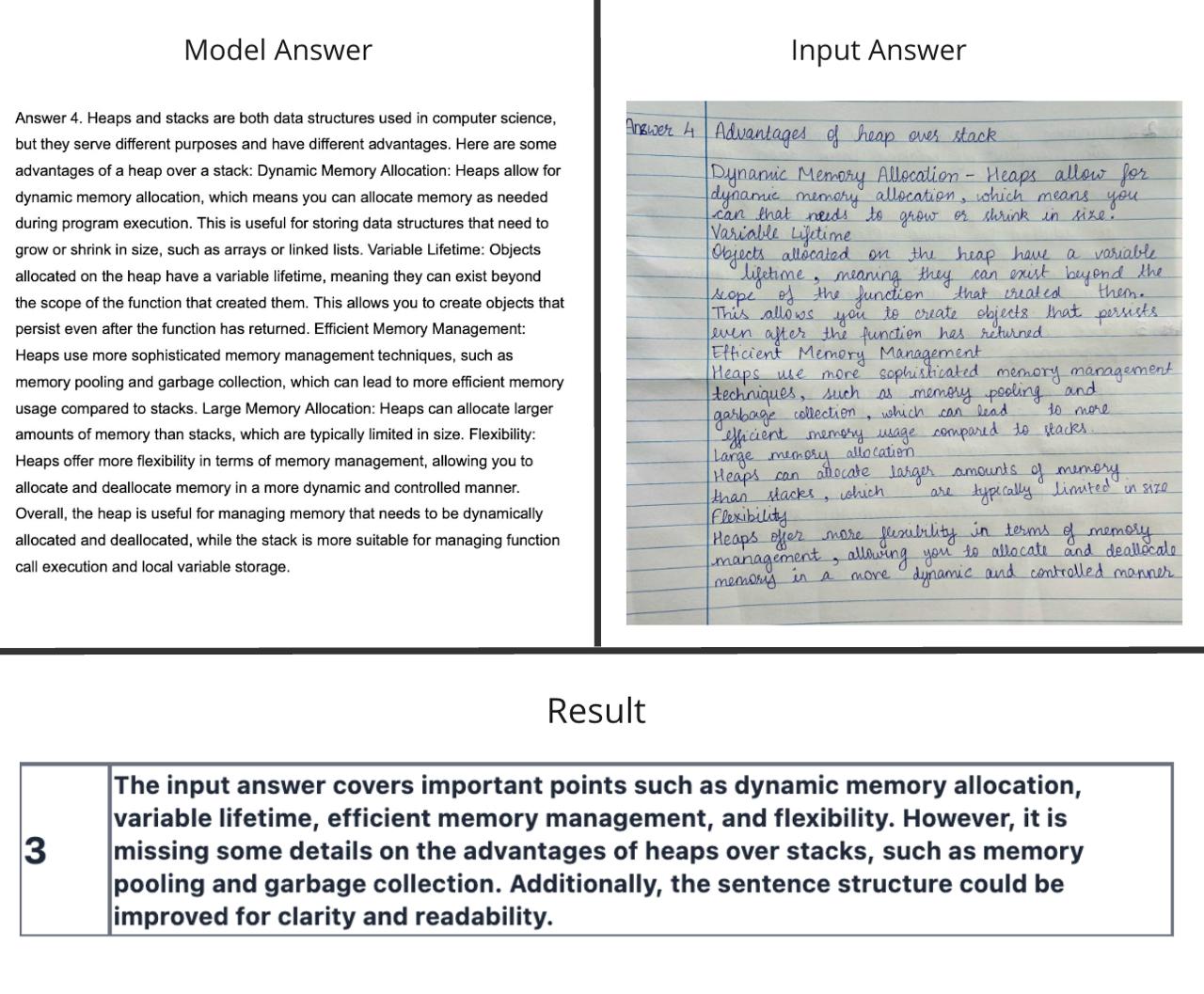}
        \caption{Test Case of Textual Evaluation}
        \label{fig:enter-label}
    \end{figure}
    \item Diagrammatic Evaluation: This involves separate evaluation of diagrams which are compared
with the sample diagrams present in the answer sheets and then
evaluated using LLMs.
    \begin{itemize}
        \item Diagrammatic evaluation was tested using a dataset containing
        both correct and incorrect diagrams.
        \item As the score is generated based on the structure of diagrams and
        labelings we used a dataset containing mistakes in either or both
        of these and tested it.
    \end{itemize}
    \begin{figure}[h]
        \centering
        \includegraphics[scale=0.18]{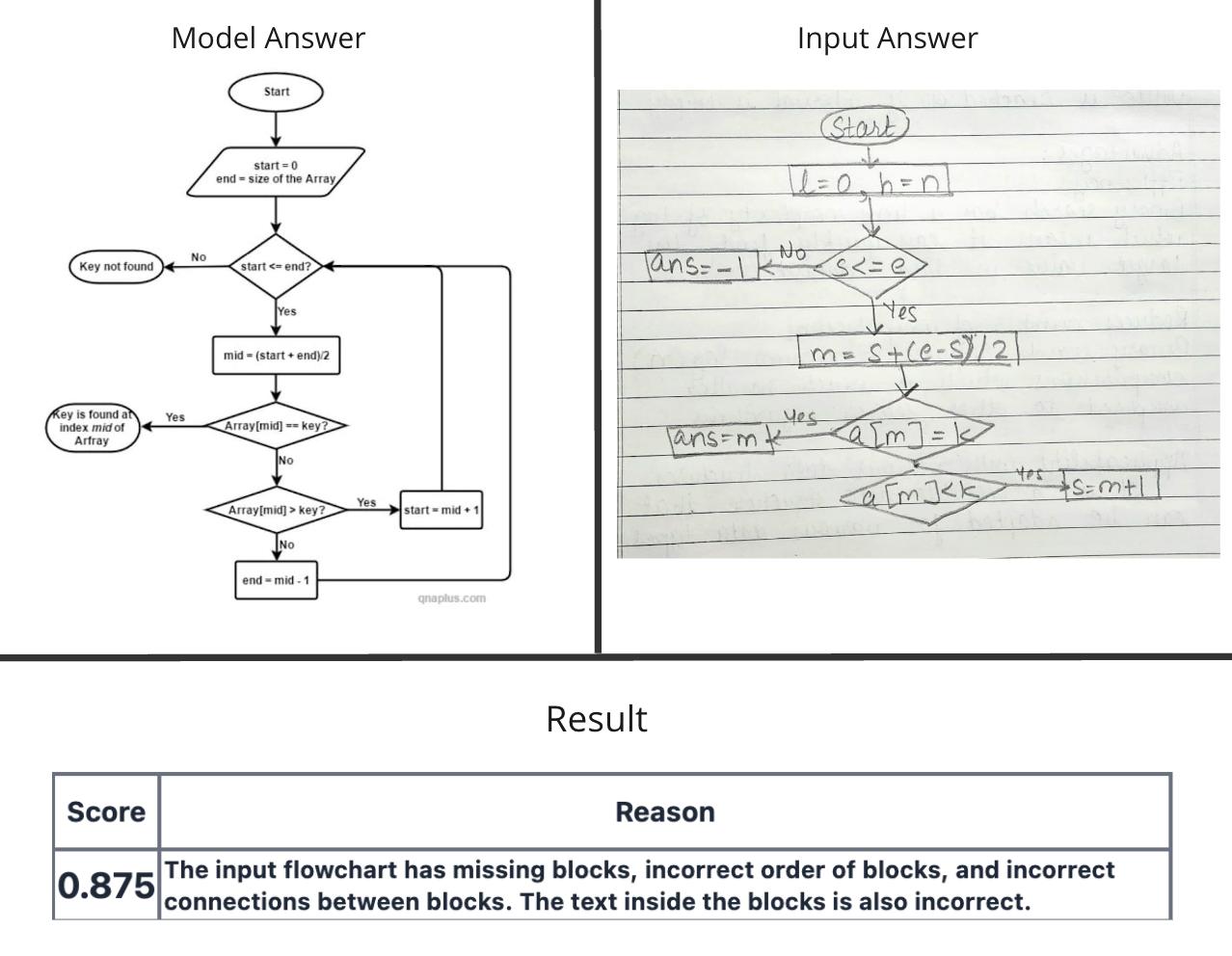}
        \caption{Test Case of Diagram Evaluation}
        \label{fig:enter-label}
    \end{figure}

\begin{figure}[h]
  \hspace*{0.5cm}  % Adjust the negative value for desired left shift (e.g., -0.5cm)
  \centering
  \includegraphics[scale=0.52]{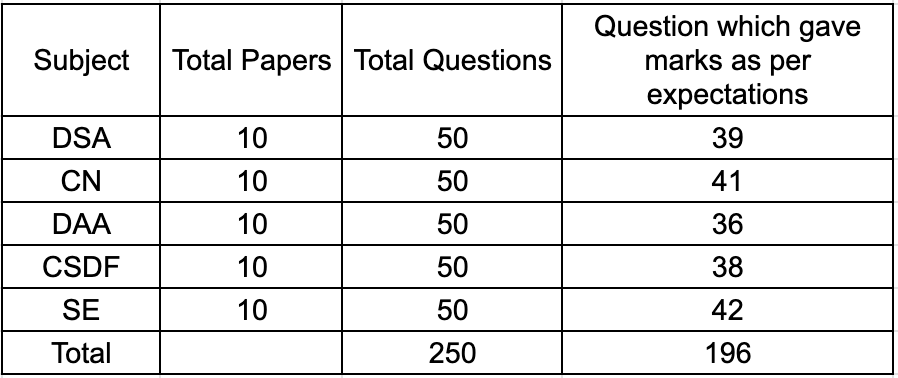}
  \caption{Paper Analysis Table}
  \label{fig:enter-label}
\end{figure}

To evaluate the model accurately, the unit test papers for students were checked for five subjects: DSA (Data Structures and Algorithms), CN (Computer Networks), DAA (Design and Analysis of Algorithms), CSDF (Computer Systems Design and Fundamentals), and SE (Software Engineering). A model answer key was created to compare with the students' answers.
Table 1 above presents the results obtained regarding students' performance in different subjects regarding the number of questions that met the expected marks. It includes the subjects evaluated, the number of papers for each subject (all being 10), the total number of questions for each subject (each having 50 questions, each question paper consisting of 5 questions), and the number of questions in each subject that gave marks according to expectations. The subjects listed are DSA (Data Structures and Algorithms), CN (Computer Networks), DAA (Design and Analysis of Algorithms), CSDF (Computer Systems Design and Fundamentals), and SE (Software Engineering). Out of 250 total questions,  196 questions gave marks as per expectations. Several factors might explain why not all questions gave perfectly accurate results. These include inaccurate or non-legible handwriting, which can lead to misinterpretation or errors in grading. Inaccurate or faintly drawn boundaries in diagrams or charts can make it difficult for graders to determine the correctness of an answer. Additionally, inadequate representation of diagrams, where diagrams are poorly drawn or not sufficiently labeled, can result in a loss of marks since they may not convey the intended context correctly.

    \begin{figure}[h]
        \centering
        \includegraphics[scale=0.18]{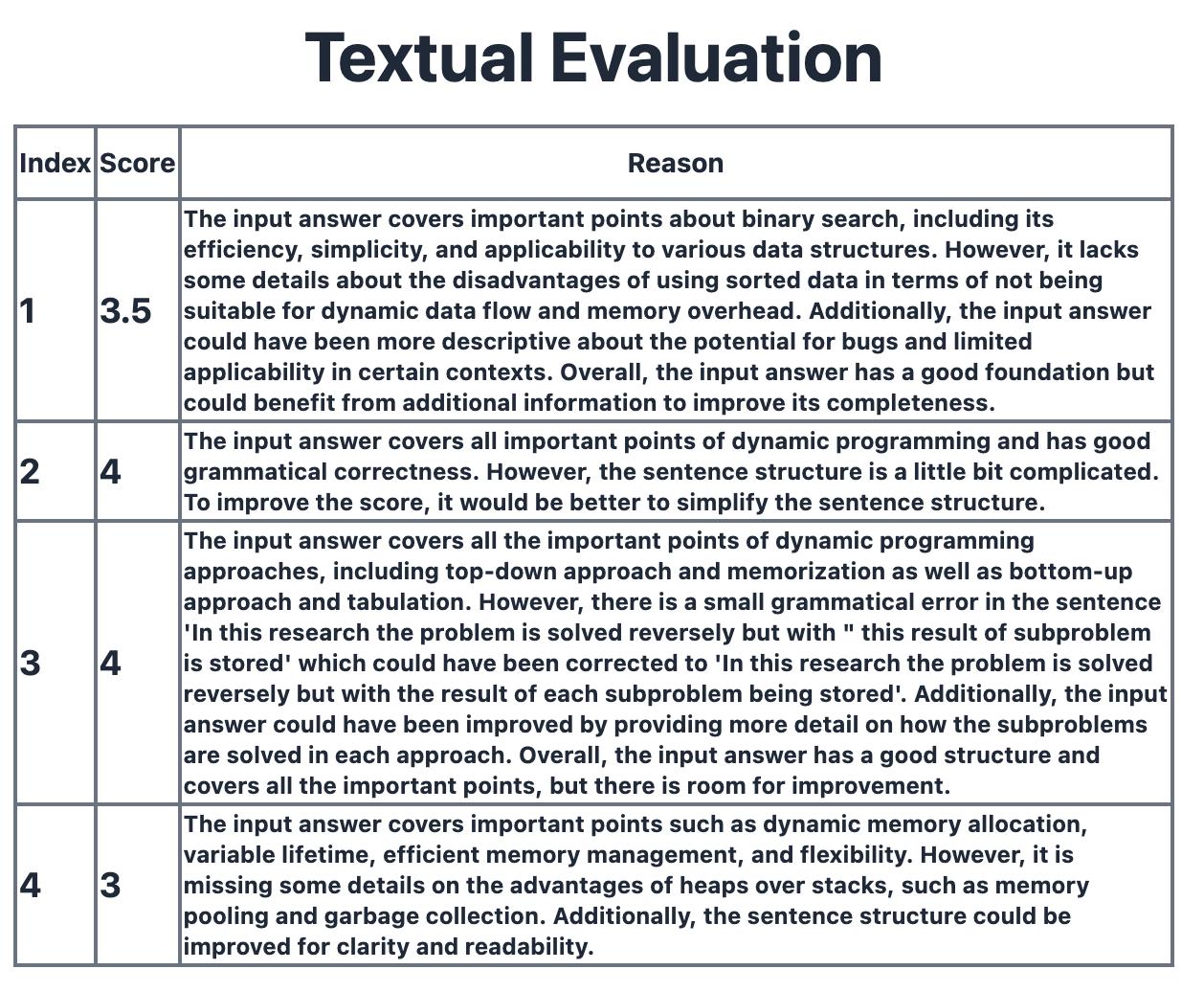}
        \includegraphics[scale=0.18]{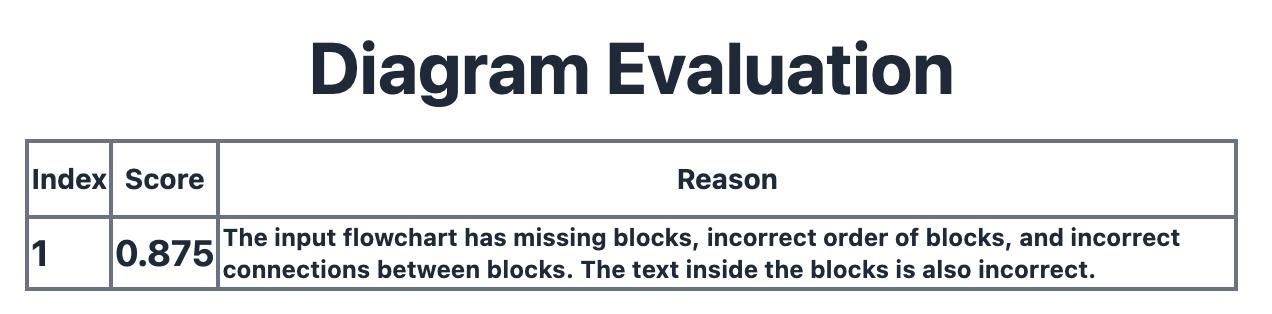}
        \caption{Results of Textual \& Diagram Evaluation}
        \label{fig:enter-label}
    \end{figure}
    
\end{enumerate}

\newpage
\section{Conclusion}
In conclusion, the innovative automated system described in this paper rep-
resents a pivotal advancement in the field of STEM education assessment.
By integrating cutting-edge technologies such as Optical Character Recognition (OCR), Natural Language Processing (NLP), and advanced diagram
understanding, our system successfully bridges the gap between textual and
visual data modalities. Furthermore, the establishment of predefined evaluation-
action thresholds underscore our commitment to objectivity and fairness in
grading, ultimately contributing to the promotion of educational equity.
This project signifies a significant milestone in the ongoing evolution
of efficient and precise automated assessment methodologies in STEM ed-
education. It envisions a future where technology not only streamlines the
evaluation process but also enriches the overall learning experience for both
educators and students. 
This proposed solution can be expanded to support other diagrams present
in the STEM domain ranging from Graph and Trees (Data Structures) to
other UML and ER Diagrams. This would require an even larger dataset.
We plan on expanding the textual evaluation part to other languages such
as local indic languages. This will increase the footfall of users and help us
in accessing further data (answer sheet images and diagram images). With
the constant development happening in LLMs, it's pretty important to keep
on integrating better Open-source LLMs as well as improved OCR systems. As we continue to push the boundaries of automated assessment, our work is poised to reshape the landscape of educational eval-
cation, making it more accessible and equitable for all.

\bibliography{main}

% \vspace{12pt}
% % \color{red}
% % IEEE conference templates contain guidance text for composing and formatting conference papers. Please ensure that all template text is removed from your conference paper prior to submission to the conference. Failure to remove the template text from your paper may result in your paper not being published.

\end{document}